# Image enhancement using the mean dynamic range maximization with logarithmic operations

Vasile Pătraşcu

Department of Informatics Technology, TAROM Company, Bucureşti-Otopeni, Romania
E-mail: vpatrascu@tarom.ro

*Abstract* - *In this paper we use a logarithmic model for gray level image enhancement. We begin with a short presentation of the model and then, we propose a new formula for the mean dynamic range. After that we present two image transforms: one performs an optimal enhancement of the mean dynamic range using the logarithmic addition, and the other does the same for positive and negative values using the logarithmic scalar multiplication. We present the comparison of the results obtained by dynamic ranges optimization with the results obtained using classical image enhancement methods like gamma correction and histogram equalization.*

## I. INTRODUCTION

Image enhancement is an important branch of image processing. Many approaches exist (contrast manipulation, histogram modification, filtering,..) that are well exposed in reference books such as [1-3]. Stockham [12] proposed an image enhancement method based on the homomorphic theory introduced by Oppenheim [6] and applied to images obtained by transmitted or reflected light. The key of this approach is to use an adapted mathematical homomorphism, which performs a transformation in order to use the classical linear mathematics and to use linear image processing techniques. Another approach exists in the general setting of logarithmic representation suited for the transmitted light imaging processes or the human visual perception. Jourlin and Pinoli introduced a mathematical framework for this kind of "non-linear" representations [4, 5]. In this paper we make a short presentation of the logarithmic model [8-11], which permits to maximize the mean dynamic range. Also we present a new formula for the mean dynamic range. The remainder of the paper is organized as follows: Section 2 introduces the addition, the real scalar multiplication, and the product for the gray levels. Similarly, the section 3 introduces the addition, the real scalar multiplication, and the product for the gray level images. Section 4 and 5 defines two optimal image transforms using the presented mathematical model. Section 6 presents experimental results and section 7 outlines the conclusions.

## II. THE REAL ALGEBRA OF THE GRAY LEVELS

We consider as the space of gray levels, the set $E = (-1, 1)$. In the set of gray levels E we will define the addition $\langle + \rangle$, the real scalar multiplication $\langle \times \rangle$ and the product $\langle \cdot \rangle$.

### A. Addition

We define the addition operation on the gray level interval by the following relation:

$$\forall v, w \in E, \quad v\langle + \rangle w = \frac{v + w}{1 + v \cdot w} \quad (1)$$

The neutral element for addition is $\theta = 0$. Each element $v \in E$ has as its opposite the element $w = -v$ and this verifies the following equation: $v\langle + \rangle w = 0$. The addition $\langle + \rangle$ is stable, associative, commutative, has a neutral element and each element has an opposite. It results that this operation establishes on $E$ a commutative group structure. We can also define the subtraction operation $\langle - \rangle$ by:

$$\forall v, w \in E, \quad v\langle - \rangle w = \frac{v - w}{1 - v \cdot w} \quad (2)$$

Using subtraction $\langle - \rangle$, we will note the opposite of $v$, with $\langle - \rangle v$.

### B. Scalar multiplication

For $\forall \lambda \in R, \forall v \in E$, we define the product between $\lambda$ and $v$ by:

$$\lambda \langle \times \rangle v = \frac{(1+v)^\lambda - (1-v)^\lambda}{(1+v)^\lambda + (1-v)^\lambda} \quad (3)$$

The two operations: addition $\langle + \rangle$ and scalar multiplication $\langle \times \rangle$ establish on $E$ a real vector space structure.

### C. The product operation

For $\forall v, w \in E$ the product $v\langle \cdot \rangle w$ is defined by the relation:

$$v\langle \cdot \rangle w = th\left(\frac{1}{4} \cdot ln\frac{1+v}{1-v} \cdot ln\frac{1+w}{1-w}\right) \quad (4)$$

We will note $v\langle \cdot \rangle v = v^{\langle 2 \rangle}$ and $v\langle \cdot \rangle v\langle \cdot \rangle v = v^{\langle 3 \rangle}$.

The neutral element for product is $u = \dfrac{e - e^{-1}}{e + e^{-1}}$ and the inverse: $v^{-1} = th\left(\dfrac{1}{arcth(v)}\right)$. The three operations, addition $\langle + \rangle$, scalar multiplication $\langle \times \rangle$ and product $\langle \cdot \rangle$ establish on $E$ a real algebra structure.

## III. THE REAL ALGEBRA OF THE GRAY LEVEL IMAGES

A gray level image is a function defined on a bi-dimensional compact $D$ from $R^2$ taking the values in the gray level space $E$. We note with $F(D,E)$ the set of gray level images defined on $D$. We can extend the operations defined on $E$ to gray level images $F(D,E)$, in a natural way:

### A. Addition

$\forall f_1, f_2 \in F(D,E), \forall (x,y) \in D$,
$(f_1 \langle + \rangle f_2)(x,y) = f_1(x,y) \langle + \rangle f_2(x,y)$

The neutral element is the function $f(x,y) = 0$ for $\forall (x,y) \in D$. The addition $\langle + \rangle$ is stable, associative, commutative, has a neutral element and each element has an opposite. As a conclusion, this operation establishes on the set $F(D,E)$ a commutative group structure.

### B. Scalar multiplication

$\forall \lambda \in R, \forall f \in F(D,E), \forall (x,y) \in D$,
$(\lambda \langle \times \rangle f)(x,y) = \lambda \langle \times \rangle f(x,y)$

The two operations, addition $\langle + \rangle$ and scalar multiplication $\langle \times \rangle$ establish on $F(D,E)$ a real vector space structure.

### C. The product operation

For $\forall f_1, f_2 \in F(D,E), \forall x,y \in D$ the product $f_1 \langle \cdot \rangle f_2$ is defined by the relation:
$(f_1 \langle \cdot \rangle f_2)(x,y) = f_1(x,y) \langle \cdot \rangle f_2(x,y)$

The three operations, addition $\langle + \rangle$, scalar multiplication $\langle \times \rangle$ and product $\langle \cdot \rangle$ establish on $F(D,E)$ a real algebra structure.

## IV. THE MEAN DYNAMIC RANGE MAXIMIZATION

### A. The dynamic range and the mean dynamic range

Let be $f \in F(D,E)$ a gray level image. Let us denote $f_i$ and $f_s$ the lower bound and the upper bound on $D$ respectively, $f_i = \inf_{(x,y) \in D}(f(x,y))$, $f_s = \sup_{(x,y) \in D}(f(x,y))$.

The dynamic range of $f$, denoted $V(f)$, is defined ([5]) as the difference: $V(f) = f_s - f_i$. $V(f)$ measures the size of the smallest contiguous interval of real numbers that encompasses all of the image values. It is based solely on two values, not on the entire value set. In practice it is not useful, due to the noise when some (few) extreme values are present. Because of that it is necessary to find a way to obtain a good estimate for the measures of the image dynamic. We propose to use for that, some statistical values of the image: the mean, the variance and the skewness. Namely we shall replace the original image with another one having only two values $v_i$ and $v_s$ so that the first 3 statistical moments are preserved ([7]). Supposing that $v_s > v_i$, we will define the mean dynamic range by: $V_m(f) = v_s - v_i$. The following algebraic system defines our values $v_i$ and $v_s$:

$$\begin{cases} q_i + q_s = 1 \\ q_i v_i + q_s v_s = m_1 \\ q_i v_i^2 + q_s v_s^2 = m_2 \\ q_i v_i^3 + q_s v_s^3 = m_3 \end{cases} \quad (5)$$

where $m_1, m_2, m_3$ are the statistical moments of order one, two and three for the image $f$.

$$m_i = \dfrac{\int_D (f(x,y))^i \, dxdy}{area(D)} \quad \text{for } i = 1,2,3$$

or $m_i = \dfrac{\sum_{(x,y) \in D}(f(x,y))^i}{card(D)}$ in the discrete case.

Using the mean $m$, the variance $\sigma$, and the skewness $s$, the system (5) becomes:

$$\begin{cases} q_i + q_s = 1 \\ q_i v_i + q_s v_s = m \\ q_i q_s (v_s - v_i)^2 = \sigma^2 \\ \sqrt{\dfrac{q_i}{q_s}} - \sqrt{\dfrac{q_s}{q_i}} = s \end{cases} \quad (6)$$

where $m = m_1$, $\sigma^2 = m_2 - (m_1)^2$ and

$s = \dfrac{m_3 - 3m_2 \cdot m_1 + 2(m_1)^3}{\sigma^3}$

Solving the system (6) we obtain:

$$\begin{cases} q_i = \dfrac{1}{2}\left(1 + \dfrac{s}{\sqrt{s^2 + 4}}\right) \\ q_s = \dfrac{1}{2}\left(1 - \dfrac{s}{\sqrt{s^2 + 4}}\right) \\ v_i = m - \sigma \sqrt{\dfrac{q_s}{q_i}} \\ v_s = m + \sigma \sqrt{\dfrac{q_i}{q_s}} \end{cases} \quad (7)$$

It is simple to prove (see [7]) that there exist the following inequalities: $f_s > v_s > v_i > f_i$. The center $v_0$ of the mean dynamic range is

$$v_0 = m + \frac{s}{2} \cdot \sigma \qquad (8)$$

When the skewness is null (i.e. $s = 0$), the histogram of the image has a symmetrical curvature, and the solution for the borders of the mean dynamic range becomes the classical one:

$$\begin{cases} v_i = m - \sigma \\ v_s = m + \sigma \end{cases} \qquad (9)$$

The classical solution (9) was used in [5], considering implicitly the skewness $s \approx 0$ that is not true for all the images.

*B. The mean dynamic range in the framework of logarithmic model.*

First, we presented the theory of the mean dynamic range in the classical algebra of the real numbers. Now we reconsider the problem in the framework of the logarithmic model. The system (6), with the logarithmic operations becomes:

$$\begin{cases} q_i + q_s = 1 \\ q_i \langle \times \rangle v_i \langle + \rangle q_s \langle \times \rangle v_s = m \\ q_i q_s \langle \times \rangle (v_s \langle - \rangle v_i)^{\langle 2 \rangle} = \sigma^{\langle 2 \rangle} \\ \sqrt{\frac{q_i}{q_s}} - \sqrt{\frac{q_s}{q_i}} = \varphi(s) \end{cases} \qquad (10)$$

and the solution of the system:

$$\begin{cases} q_i = \frac{1}{2}\left(1 + \frac{\varphi(s)}{\sqrt{\varphi^2(s)+4}}\right) \\ q_s = \frac{1}{2}\left(1 - \frac{\varphi(s)}{\sqrt{\varphi^2(s)+4}}\right) \\ v_i = m \langle - \rangle \sqrt{\frac{q_s}{q_i}} \langle \times \rangle \sigma \\ v_s = m \langle + \rangle \sqrt{\frac{q_i}{q_s}} \langle \times \rangle \sigma \end{cases} \qquad (11)$$

where $\varphi : E \to R$, $\varphi(v) = arcth(v)$, $\forall v \in E$ (see [10], [11]). The center of the mean dynamic range is:

$$v_0 = m \langle + \rangle \frac{\varphi(s)}{2} \langle \times \rangle \sigma \qquad (12)$$

In the discrete case the mean, the variance and the skewness are computed with the following relations:

$$m = \frac{1}{card(D)} \langle \times \rangle \left( \underset{(x,y) \in D}{\langle + \rangle} f(x,y) \right)$$

$$\sigma^{\langle 2 \rangle} = \frac{1}{card(D)} \langle \times \rangle \left( \underset{(x,y) \in D}{\langle + \rangle} (f(x,y) \langle - \rangle m)^{\langle 2 \rangle} \right)$$

$$\varphi(s) = \frac{1}{card(D)} \sum_{(x,y) \in D} \left( \frac{\varphi(f) - \varphi(m)}{\varphi(\sigma)} \right)^3$$

*C. The maximization of the mean dynamic range using the logarithmic addition*

Let be $f \in F(D,E)$ an image and $w \in E$ a gray level value. A $w$ - translation of image $f$ is defined as $f \langle + \rangle w$ and consequently the mean dynamic range of translation is $V_m(f \langle + \rangle w) = v_s \langle + \rangle w - v_i \langle + \rangle w$. The class $(f \langle + \rangle w)_{w \in E}$ of the translations associated with an image $f$ appears naturally as the set of reference where the solution is to be found. The optimization problem is to find the translation with the larger mean dynamic range. There exists a unique gray level, denoted $w_0$, such that the image $f \langle + \rangle w_0$ presents the maximal dynamic range in the $(f \langle + \rangle w)_{w \in E}$ class, namely $V_m(f \langle + \rangle w_0) = \underset{w \in E}{max}(V_m(f \langle + \rangle w))$ where $w_0$ is explicitly defined by:

$$w_0 = -\left( m \langle + \rangle \frac{\varphi(s)}{2} \langle \times \rangle \sigma \right) \qquad (13)$$

Thus the translation $T$ that performs the optimal enhancement of the mean dynamic range is defined by: $T(f) = f \langle + \rangle w_0$. We shall proceed to prove (13). Let be the function $h : [-1,1] \to [0,2]$ by putting $h(w) = v_s \langle + \rangle w - v_i \langle + \rangle w$, namely

$$h(w) = \frac{v_s + w}{1 + w v_s} - \frac{v_i + w}{1 + w v_i}$$

The first derivative is:

$$h'(w) = \frac{1 - v_s^2}{(1 + w v_s)^2} - \frac{1 - v_i^2}{(1 + w v_i)^2} .$$

We must observe the change of the sign between the values $h'(-1)$ and $h'(1)$. Thus:

$$h'(-1) = \frac{2(v_s - v_i)}{(1 - v_s)(1 - v_i)} > 0$$

and $\quad h'(1) = \frac{2(v_i - v_s)}{(1 - v_s)(1 - v_i)} < 0 .$

Solving the equation $h'(w) = 0$ yields a unique solution $w_0 \in E$ such that $w_0 = \dfrac{\sqrt{1 - v_s^2} - \sqrt{1 - v_i^2}}{v_s \sqrt{1 - v_i^2} - v_i \sqrt{1 - v_s^2}}$. It is easier to prove that the solution $w_0$ verifies the relations:

$$\left(\frac{1 + w_0}{1 - w_0}\right)^2 = \left(\frac{1 - v_i}{1 + v_i}\right)\left(\frac{1 - v_s}{1 + v_s}\right)$$

and $\quad w_0 = -\left(\frac{1}{2} \langle \times \rangle (v_i \langle + \rangle v_s)\right)$

Because the second derivative

$$h^{''}(w_0) = -2\frac{\sqrt{(1-v_s^2)(1-v_i^2)}}{(1+w_0 v_s)^2 (1+w_0 v_i)^2}(v_s - v_i) < 0$$, it can state:

$h(w_0)$ is the maximal value of the function $h$ on the gray level set $E$.

We must observe that after an optimal translation one obtains an enhancement for the image brightness.

## V. THE POSITIVE AND NEGATIVE DYNAMIC RANGE MAXIMIZATION

### A. The positive mean dynamic range

Let be an image $f \in F(D,E)$. We consider the following subset of the support $D$: $D_p = ((x,y) \in D | f(x,y) > 0)$. Let be $f_p$ the function defined on $D_p$ preserving the values of $f$. Namely $\forall (x,y) \in D_p$, $f_p : D_p \to E^+$ with $f_p(x,y) = f(x,y)$. The mean dynamic range $V_m(f_p)$ is the mean dynamic range for the positive values of $f$. Let be $m_p, \sigma_p, s_p$ the mean, the variance and the skewness of the function $f_p$. From the relations (11), we compute $q_{pi}, q_{ps}, v_{pi}, v_{ps}$ for the function $f_p$. Thus it result the mean dynamic range: $V_m(f_p) = v_{ps} - v_{pi}$

### B. The negative mean dynamic range

Let be an image $f \in F(D,E)$. We consider the following subset of the support $D$: $D_n = ((x,y) \in D | f(x,y) < 0)$. Let be $f_n$ the function defined on $D_n$ preserving the values of $f$. Namely $\forall (x,y) \in D_n$, $f_n : D_n \to E^-$ with $f_n(x,y) = f(x,y)$. The mean dynamic range $V_m(f_n)$ is the mean dynamic range for the negative values of $f$. Let be $m_n, \sigma_n, s_n$ the mean, the variance and the skewness of the function $f_n$. From the relations (11), we compute $q_{ni}, q_{ns}, v_{ni}, v_{ns}$ for the function $f_n$. Then it result the mean dynamic range $V_m(f_n) = v_{ns} - v_{ni}$

### C. The mean dynamic range optimization for positive values

A $\lambda$-homothetic of $f_p$ is defined as $\lambda \langle \times \rangle f_p$ and consequently the mean dynamic range of positive homothetic is $V_m(\lambda \langle \times \rangle f_p) = \lambda \langle \times \rangle v_{ps} - \lambda \langle \times \rangle v_{pi}$ where $\lambda$ is a positive real number. The class $(\lambda \langle \times \rangle f_p)_{\lambda > 0}$ of strictly positive homothetic associated with an $f_p$ appears naturally as the set of reference where the solution is to be found. The optimization problem is to find the positive homothetic with the larger mean dynamic range, supposing that $\sigma_p \neq 0$.

Under the previous conditions, there exists a unique strictly positive real number, denoted $\lambda_p$, such that the function $\lambda_p \langle \times \rangle f_p$ presents the maximal mean dynamic range in the $(\lambda \langle \times \rangle f_p)_{\lambda > 0}$ class, namely:

$$V_m(\lambda_p \langle \times \rangle f_p) = \max_{\lambda > 0}(V_m(\lambda \langle \times \rangle f_p)).$$

We shall proceed to compute $\lambda_p$. Define the function $h : (0, \infty) \to [0,1]$ by putting $h(\lambda) = \lambda \langle \times \rangle v_{ps} - \lambda \langle \times \rangle v_{pi}$.

Let be: $u_{ps} = \frac{1 - v_{ps}}{1 + v_{ps}}$ and $u_{pi} = \frac{1 - v_{pi}}{1 + v_{pi}}$.

There are the inequalities: $1 > u_{pi} > u_{ps} > 0$. With these notation $h(\lambda)$ becomes:

$$h(\lambda) = 2\left(\frac{1}{1 + u_{ps}^\lambda} - \frac{1}{1 + u_{pi}^\lambda}\right)$$

The first derivative of $h$ is:

$$h'(\lambda) = -2\left(\frac{u_{ps}^\lambda \ln(u_{ps})}{(1 + u_{ps}^\lambda)^2} - \frac{u_{pi}^\lambda \ln(u_{pi})}{(1 + u_{pi}^\lambda)^2}\right)$$

The equation $h'(\lambda) = 0$ has the following form

$$\lambda = \frac{\ln\left(\frac{\ln(u_{pi})}{\ln(u_{ps})}\right) + 2\ln\left(\frac{1 + u_{ps}^\lambda}{1 + u_{pi}^\lambda}\right)}{\ln\left(\frac{u_{ps}}{u_{pi}}\right)} \qquad (14)$$

The equation (14) supply us the recurrence relation to compute the solution. There are true the inequalities

$$1 > \frac{1 + u_{ps}^\lambda}{1 + u_{pi}^\lambda} > \frac{1}{2} \qquad (15)$$

From (15) it results that

$$\lambda \in \left(\frac{\ln\left(\frac{\ln(u_{pi})}{\ln(u_{ps})}\right) + 2\ln\left(\frac{1}{2}\right)}{\ln\left(\frac{u_{ps}}{u_{pi}}\right)}, \frac{\ln\left(\frac{\ln(u_{pi})}{\ln(u_{ps})}\right)}{\ln\left(\frac{u_{ps}}{u_{pi}}\right)}\right)$$

Because of that the relation (14) generates a bounded string. We will start the string computation with value

$$\lambda_0 = \frac{\ln\left(\frac{\ln(u_{pi})}{\ln(u_{ps})}\right) + 2\ln\left(\frac{1 + u_{ps}}{1 + u_{pi}}\right)}{\ln\left(\frac{u_{ps}}{u_{pi}}\right)}.$$

Two or three iterations are enough to obtain a good precision. Let be $\lambda_p$ the solution of the equation (14). The second derivative $h''(\lambda)$ for $\lambda = \lambda_p$ has the following value:

$$h''(\lambda_p) = -2\frac{u_{ps}^{\lambda_p} \ln(u_{ps})}{(1+u_{ps}^{\lambda_p})^2}\left(\frac{1-u_{ps}^{\lambda_p}}{1+u_{ps}^{\lambda_p}}\ln(u_{ps}) - \frac{1-u_{pi}^{\lambda_p}}{1+u_{pi}^{\lambda_p}}\ln(u_{pi})\right)$$

Because $h''(\lambda_p) < 0$, it can state: $h(\lambda_p)$ is the maximal value of function $h$.

### D. The mean dynamic range optimization for negative values

The optimization problem of the mean dynamic range for negative values is similarly to that for positive values, as described above. As a consequence, it will only be made a short presentation for the solving of the negative values problem. A $\lambda$ - homothetic of $f_n$ is defined as $\lambda\langle\times\rangle f_n$ and consequently the mean dynamic range of positive homothetic is $V_m(\lambda\langle\times\rangle f_n) = \lambda\langle\times\rangle v_{ns} - \lambda\langle\times\rangle v_{ni}$ where $\lambda$ is a positive real number. Supposing that $\sigma_n \neq 0$, there exists a unique strictly positive real number, denoted $\lambda_n$, such that the function $\lambda_n\langle\times\rangle f_n$ presents the maximal mean dynamic range, namely $V_m(\lambda_n\langle\times\rangle f_n) = \max_{\lambda>0}(V_m(\lambda\langle\times\rangle f_n))$. Define the function $h:(0,\infty) \to [0,1]$ by putting $h(\lambda) = \lambda\langle\times\rangle v_{ns} - \lambda\langle\times\rangle v_{ni}$. Let be: $u_{ns} = \frac{1-v_{ns}}{1+v_{ns}}$ and $u_{ni} = \frac{1-v_{ni}}{1+v_{ni}}$. The equation $h'(\lambda) = 0$ has the following form

$$\lambda = \frac{\ln\left(\frac{\ln(u_{ni})}{\ln(u_{ns})}\right) + 2\ln\left(\frac{1+u_{ns}^{\lambda}}{1+u_{ni}^{\lambda}}\right)}{\ln\left(\frac{u_{ns}}{u_{ni}}\right)} \quad (16)$$

The equation (16) supply us the recurrence relation to compute the solution. We will start the string computation with value $\lambda_0 = \frac{\ln\left(\frac{\ln(u_{ni})}{\ln(u_{ns})}\right) + 2\ln\left(\frac{1+u_{ns}}{1+u_{ni}}\right)}{\ln\left(\frac{u_{ns}}{u_{ni}}\right)}$. Two or three iterations are enough to obtain a good precision. If $\lambda_n$ is the solution of the equation (16) then $h(\lambda_n)$ is the maximal value of function $h$. We must also observe that after an optimal homothetic of positive and negative gray level values one obtains an enhancement for the image contrast.

## VI. EXPERIMENTAL RESULTS

To exemplify, two images were picked out: one dark ("fe1g") in Fig.1a and one bright ("cells") in Fig.2a. The Figs. 1d and 2d present the result of the enhancement by the proposed logarithmic approach. In each figure, the results of the application of some classical enhancement techniques (histogram equalization in Figs. 1b and 2b, gamma correction in Figs. 1c and 2c) on the same test images are also shown. Based on these typical examples, we claim that the proposed enhancement technique provide significant improvements compared to the classical automatic (or adaptive) gray level image enhancement methods.

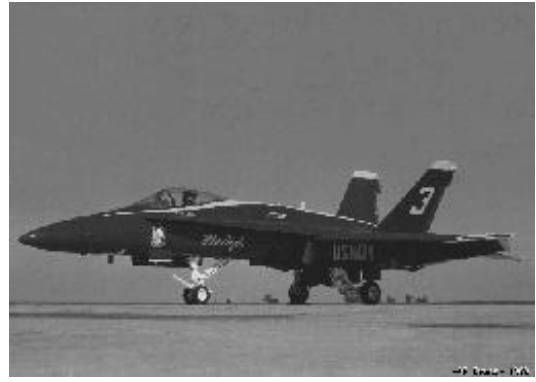

(a) Original image "fe1g"

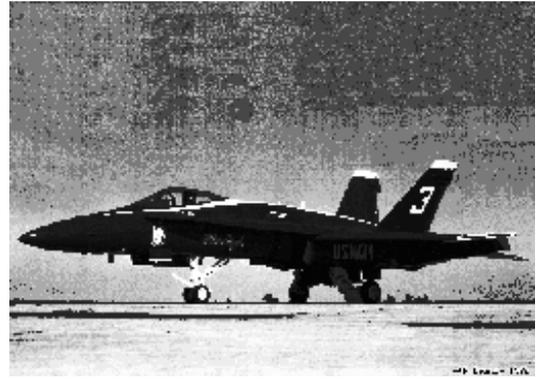

(b) Histogram equalization

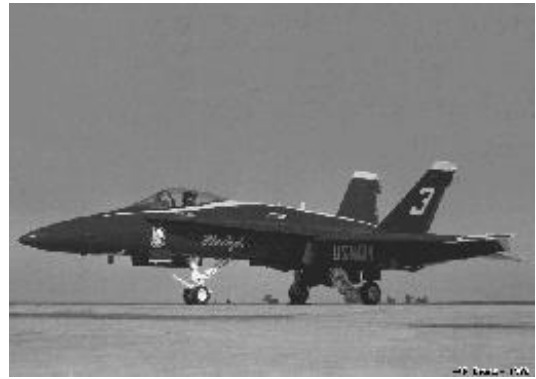

(c) Gamma correction

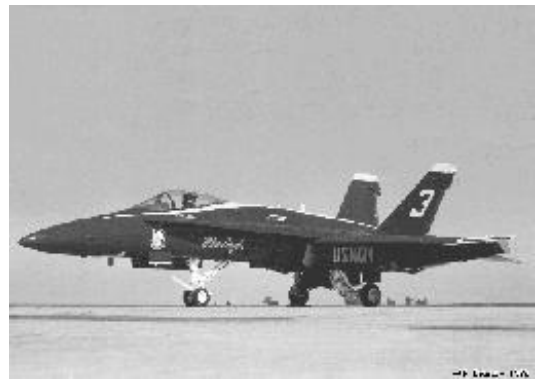

(d) Mean dynamic range maximization

Fig. 1

## VII. CONCLUSION

The image enhancement has been and is always an important field of the research where publications of new techniques and methods take large places in the literature. A great number of enhancement methods exists because each application is specific and needs an adapted method [2]. So, the physical nature of images to be processed is of central importance and the need of an adequate image mathematic model appears clearly as a necessity [5]. In this paper we have presented two image enhancement transformations, using the logarithmic operations: addition and scalar multiplication. Also we presented a new formula for the mean dynamic range. The tests show that the proposed techniques allow the automatic correction of the illumination problems that occur during the acquisition process, yielding a better result than some classical image enhancement methods, like the histogram equalization or gamma correction, for gray level images. We can say that the proposed methods are another strong argument for the rich potential of the logarithmic image processing models.

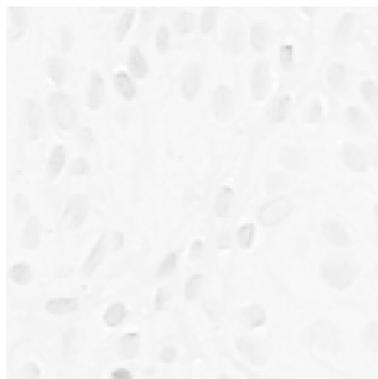

(a) Original image "cells"

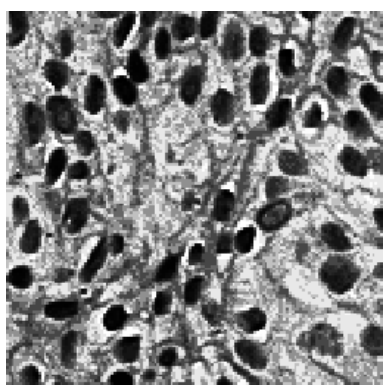

(b) Histogram equalization

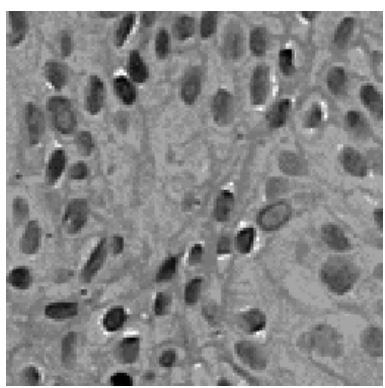

(c) Gamma correction

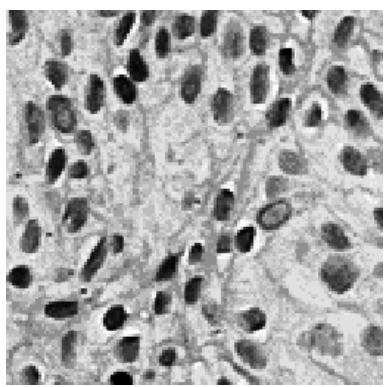

(d) Mean dynamic range maximization

Fig. 2